# Driving-Policy Adaptive Safeguard for Autonomous Vehicles Using Reinforcement Learning

Zhong Cao, Shaobing Xu, Songan Zhang, Huei Peng, Diange Yang

*Abstract*—Safeguard functions such as those provided by advanced emergency braking (AEB) can provide another layer of safety for autonomous vehicles (AV). A smart safeguard function should adapt the activation conditions to the driving policy, to avoid unnecessary interventions as well as improve vehicle safety. This paper proposes a driving-policy adaptive safeguard (DPAS) design, including a collision avoidance strategy and an activation function. The collision avoidance strategy is designed in a reinforcement learning framework, obtained by Monte-Carlo Tree Search (MCTS). It can learn from past collisions and manipulate both braking and steering in stochastic traffics. The driving-policy adaptive activation function should dynamically assess current driving policy risk and kick in when an urgent threat is detected. To generate this activation function, MCTS's exploration and rollout modules are designed to fully evaluate the AV's current driving policy, and then explore other safer actions. In this study, the DPAS is validated with two typical highway-driving policies. The results are obtained through and 90,000 times in the stochastic and aggressive simulated traffic. The results are calibrated by naturalistic driving data and show that the proposed safeguard reduces the collision rate significantly without introducing more interventions, compared with the state-based benchmark safeguards. In summary, the proposed safeguard leverages the learning-based method in stochastic and emergent scenarios and imposes minimal influence on the driving policy.

## I. INTRODUCTION

AUTONOMOUS technologies have great potential to improve vehicle safety and mobility [1]. Safeguard systems [2] have been proposed, independent of the higher-level co-pilot or chauffer functions to avoid collisions, which can be implemented on both human-driven vehicles and autonomous vehicles. National Highway Traffic Safety Administration (NHTSA) reported that safeguard functions such as advanced emergency braking (AEB [3]) can prevent 28,000 collisions and 12,000 injuries annually [4]. For autonomous vehicles, they are usually driven by different driving policies (e.g., car-following and eco-driving) and consider various performance targets, e.g., safety, efficiency, mobility and comfort [5] [6]. These higher-level co-pilot or chauffer functions may not be absolutely safe, e.g., due to the uncertain motions of other road users [7]. The safeguards are designed to take last-second action to avoid a collision or reduce its severity, which usually have two parts, i) determining the activation timing and ii) controlling the vehicle to avoid the collision.

An ideal safeguard for high-speed driving should ensure a high level of safety while maintaining minimal intervention in stochastic traffic. The challenges are i) generating the collision avoidance strategies under strong traffic uncertainty and using both braking and steering to improve the driving safety; ii) designing a driving-policy adaptive activation function. The activation function should avoid unnecessary interventions when the driving policy can keep safety, or activate the safeguard in advance if the driving policy cannot. To achieve this goal, this paper designs a driving-policy adaptive safeguard (DPAS) algorithm on the highway. The performance metrics of the safeguard include both collision rate and intervention rate.

The main contribution of this paper includes:

1) a collision avoidance strategy on highway using reinforcement learning method obtained by Monte-Carlo tree search. The safeguard can manipulate both braking and steering when an urgent threat is detected under strong traffic uncertainty and high-speed scenarios.

2) a driving-policy adaptive activation function. The activation function can estimate the driving policy's performance and activate the safeguard only when the policy will perform poorly. The MCTS designs the rollout and exploration policy to fully evaluate the driving policy, and only explore when it may get collisions. Such a setting can adapt the activation conditions to the driving policies, which can avoid unnecessary interventions as well as improve vehicle safety.

3) for training and testing the safeguard, a simulation platform is designed based on the accelerated evaluation method [24]. An important-sampling based approach calibrates the simulation results by naturalistic driving data. It helps reduce simulation time significantly.

The remainder of this paper is organized as follows: Section II summarizes the related literature. The problem of the proposed safeguard system is described in Section III. Section IV establishes the DPAS framework. The traffic simulation

Z. Cao is with the Department of Automotive Engineering, Tsinghua University, Beijing, China, 100084, caoc15@mails.tsinghua.edu.cn, and the Department of Mechanical Engineering, University of Michigan, Ann Arbor, MI, 48105, zhcao@umich.edu

H. Peng, S. Xu, S. Zhang are with the Department of Mechanical Engineering, University of Michigan, Ann Arbor, MI, 48109. (email: hpeng, xushao, songanz@umich.edu, respectively)

D. Yang is with the Department of Automotive Engineering, Tsinghua University, Beijing, China, 100084, ydg@tsinghua.edu.cn

platform is described in Section V. The design of the reinforcement learning algorithm is explained in Section VI. The simulation results are presented and analyzed in Section VII. Finally, Section VIII concludes the paper.

## II. LITERATURE REVIEW

### A. Safeguard Activation Conditions

Most safeguards' activation conditions are determined by the current driving state, e.g., car-following range or range rate. Intel and Mobileye proposed a conservative activation condition called responsibility sensitive safety (RSS) [9]. It assumes the surrounding vehicles will operate aggressively, e.g., the front vehicle brakes suddenly or vehicles from neighbor lanes may abruptly cut in. The safeguard will be activated when the ego vehicle will enter the dangerous states under these assumptions. Therefore, this strategy requires maintaining long car-following distance, which differs from natural driving and may lead to more than frequent intervention.

Some other activation conditions are designed less conservatively [3][8] using factors such as Time-to-collision (TTC), Time headway (THW), and car-following distance. The activation threshold can be tuned manually [10], based on human driving data [3], or based on prediction [11]. These safeguards are hard to set the proper activation threshold in stochastic traffic.

The driving-state based activation conditions cannot analyze and utilize the AV driving policies. It may cause lower safety if the safeguard cannot be activated in advance when the driving policy performs poorly, or unnecessary and frequent interventions when the driving policy can avoid the collisions. The proposed DPAS aims to adapt the activation conditions to the driving policy, which can avoid unnecessary interventions as well as improve vehicle safety.

### B. Collision Avoidance Strategies

The challenges of the collision avoidance strategies are the long planning horizon and strong traffic uncertainty, especially in high-speed driving scenarios.

Some of collision avoidance strategies [8] use braking only for simplicity, while in theory, both braking and steering can be used. Using both braking and steering is more complex, considering more vehicles in 2D rather than 1D. Two typical safeguards are based on worst cases or predictions. The worst-case based safeguard considers the worst cases of the environment and avoids collisions based on the reachable set [12], Hamilton-Jacobi-Bellman equation or barrier functions [13]. Some other reported safeguards are designed based on surrounding vehicle predictions and solved by rule/optimization/sampled strategies. These safeguards balance between mobility, ride comfort and safety. For instance, J. Funke et al. [14] solved this problem using the model predictive control (MPC) technique. Similarly, RRT [15], fuzzy logic [16], and motion primitive [17] methods could be used. However, a prediction-based safeguard requires some assumptions; for example, the surrounding vehicles always follow a specific driver model or the vehicle speed can be controlled instantaneously [19]. They may not be able to deal with highly stochastic traffic.

Recently, learning-based path planning algorithms [20] were developed in autonomous vehicles. Usually, these algorithms are used alone instead of being a safeguard [22]. Some RL based safeguards [23] are based on the observed state, may not fully utilize the driving policies, which may cause frequent interventions.

## III. PROBLEM STATEMENT

The proposed DPAS is designed for the scenario shown in Fig. 1: *An AV driving on the highway is being controlled by given driving policies (e.g., car-following and lane-changing). Its surrounding traffic contains uncertainty in other vehicles' driving behaviors and their interactions. The stochastic traffic may lead to unexpected emergency cases (e.g. aggressive cut-in). Therefore, a driving-policy adaptive safeguard should detect the urgent threat according to the surrounding environment and the driving policy ability, and modify the longitudinal and lateral motions to avoid collisions.*

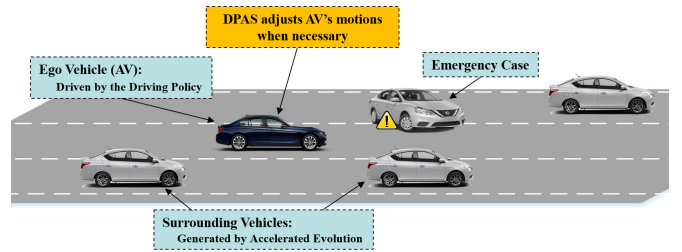

Fig. 1. The highway driving scenario studied in our DPAS problem

Fig. 1 shows a DPAS application scenario. The simulated traffic environment has the following features, and Section V will describe the method to generate this stochastic simulator.

*a) Uncertain driving behaviors:* Each surrounding vehicle has a randomly generated driving strategy, represented by different parameter values of the driver models. These driving strategies are unobserved by the AV. Moreover, surrounding vehicles may deviate from their driving strategies during driving to add further randomness.

*b) Interaction between vehicles:* A surrounding vehicle's motions can be affected by both the ego AV and other vehicles.

*c) Aggressive traffic:* The traffic environment is designed to accelerate the exposure to higher-risk driving conditions, i.e., the driver model parameters are sampled to represent risky drivers.

*d) Calibrated by naturalistic driving data:* The final performance metrics are calibrated by naturalistic driving data [37] after simulation, using important sampling.

Such traffic setting makes it hard to accurately predict surrounding vehicles, thus more difficult to detect the activation timing and then plan the escapement trajectories.

The DPAS is designed under the following assumptions: 1) the states of the surrounding vehicles within a finite range (+/- 100 m) are observed, which may contain noise from sensors measurements (e.g., light detection and ranging (LIDAR) and camera [25]). 2) The ego AV's control module (e.g., PID control or preview control [26]) can follow a planned trajectory, which was designed to respect the vehicle dynamic constraints.

The Gipps car-following model [27] and a human-like driver model (described in Section V-B) are taken as two examples of driving policies for the ego AV. The conservative safeguards, i.e., RSS considering brake only and reachable-set based safeguard considering both brake and steering, are used as benchmarks, described in Section V-C.

## IV. ARCHITECTURE OF THE DPAS

The proposed DPAS framework uses reinforcement learning and involves in the current driving policy for the activation function design, shown in Figure 2.

In the following, we will first describe the major elements of designing reinforcement learning, including the system states, actions, and rewards. Then, introduce the activation function design considering the driving policy.

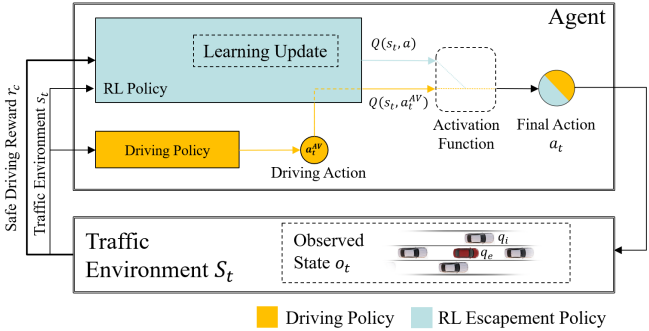

Fig. 2. The DPAS framework using reinforcement learning. The blue block represents the reinforcement learning framework and the yellow block represents the driving policy.

### A. Reinforcement Learning

*1) State Space:* The state space $\mathcal{S}$ is defined as follows:

$$s = (q_e, \{q_1, q_2 \ldots, q_n\}, \{\theta_1, \theta_2 \ldots, \theta_n\}) \in \mathcal{S} \quad (1)$$

where $q_e = (x_e, y_e, \dot{x}_e, \dot{y}_e)$ represents the ego AV's physical state. The surrounding vehicles' longitudinal/lateral positions and speeds are $q_i = (x_i, y_i, \dot{x}_i, \dot{y}_i)$. The parameter $\theta_i$ is an internal factor, denoting the estimated driver strategy of the $i^{th}$ surrounding vehicle, which is defined and estimated in Section VI-A. It is necessary because different driver characteristics should bring different decisions under the same kinematic states.

*2) Action Space:* The action $a$ in the action space $\mathcal{A}$ is defined as follows:

$$a = (\ddot{x}_e, y_{lc}) \in \mathcal{A} \quad (2)$$

where $\ddot{x}_e$ denotes the longitudinal acceleration. The value of $y_{lc}$ belongs to the set $\{0, 1/6, -1/6\}$, denoting lane keeping (0), slight right (-1/6), slight left (+1/6) lane changing. The value 1/6 means that a complete lane change requires six consecutive lane-change decisions in the same direction. Such settings are required to accommodate the nature of lane change operations.

The actions in Eq. (2) indicate the desired position and velocity of the ego vehicle after a time step $\Delta t$ (set as 0.75s), calculated as follows:

$$x_e(t + \Delta t) = x_e(t) + \dot{x}_e(t)\Delta t + 0.5\ddot{x}_e\Delta t^2$$
$$\dot{x}_e(t + \Delta t) = \dot{x}_e(t) + \ddot{x}_e\Delta t \quad (3)$$
$$y_e(t + \Delta t) = y_e(t) + y_{lc}w$$

where $w$ denotes the lane width. The 0.75s is chosen according to the driver reaction time (about 0.7s-1.5s) and a trade-off between smooth driving and faster calculation. Other reasonable value also works in our framework.

For simplicity, we pre-select a subset in the action space as candidate actions. These candidate actions consist of two parts: driving action $a_t^{AV}$ and collision avoidance action set $\{a_t^{ca}\}$.

$$a \in \{a_t^{AV}, \{a_t^{ca}\}\} \quad (4)$$

where driving action $a_t^{AV}$ is generated by the driving policy. Collision avoidance actions $a_t^{ca}$ are used when RL needs to take actions to intervene. In the collision avoidance action set, $\ddot{x}_e$ is selected from hard braking, light braking, coasting, and light acceleration.

Every action in the action space indicates a smoothness trajectory, which the AV will track on the road. In this work, given an action, the trajectory is generated with a fifth-order polynomial curve, constrained by $(x_e(t_c), \dot{x}_e(t_c))$ and $(x_e(t_c + \Delta t), \dot{x}_e(t_c + \Delta t))$ from Eq. (3). Furthermore, the actions causing the ego vehicle to drive too slow or too fast, or change to an unavailable lane, are discarded.

*3) Reward:* The reward $r$ is the safe-driving reward. The reward is earned when AV is driving on the road safely, defined as:

$$r = r_c, \text{ while driving}$$
$$r = 0, \text{ after collision} \quad (5)$$

where $r_c \in \mathbf{R}^+$ is a positive constant.

According to the reward function, the performance of a training policy is only decided by collisions. AV can set other tasks in the driving policies and use the safeguard to keep safety in the stochastic environment.

### B. Activation Function Design

The activation function design uses the state-action value $Q_\pi(s, a)$, which is estimated in the reinforcement learning policy update process. The Q value function is defined as:

$$Q_\pi(s, a) := \mathbb{E}_\pi \left[ \sum_{t=h}^{h+H-1} \gamma^{t-h} r_t \Big| s_h = s, a_h = a \right] \quad (6)$$

where $\mathbb{E}[\cdot]$ denotes the expectation with respect to the environment transition distribution. $r_t$ denotes the reward at time $t$. $H$ denotes the time horizon. $\gamma$ denotes the discount factor. $\pi$ denotes the policy, which could be the driving policy and the collision avoidance policy.

In Eq. (6), $Q_\pi(s,a)$ means the discount rewards expectation in the following $H$ time steps. According to the reward setting, less value $Q_\pi(s,a)$ means more risky policy. Therefore, $Q_\pi(s,a)$ represents the risk level the AV uses the policy $\pi$ to drive in the following $H$ time steps. Compared with the state-based activation conditions, using $Q_\pi(s,a)$ can take the driving policy ability into account, to avoid unnecessary interventions.

The activation function is as follows:

$$\mathcal{C}(s,\pi^{AV}) = Q(s,a) - Q_{\pi^{AV}}(s,a) - c_{\text{thres}} \quad (7)$$

where $\mathcal{C}(s,\pi^{AV})$ denotes the activation function. $Q_{\pi^{AV}}(s,a)$ denotes the value using the current driving policy in the following $H$ time steps. $Q(s,a)$ means the value using other policies. $c_{\text{thres}}$ is the manually designed threshold, discussed in Section VI-C. The safeguard will be activated when $\mathcal{C}(s,\pi^{AV}) > 0$.

The activation function requires a fixed policy value $Q_{\pi^{AV}}(s,a)$, while the conventional RL framework will always update the policy for better performance. Therefore, the paper adjusts the MCTS framework for the activation function, which can fully estimate $Q_{\pi^{AV}}(s,a)$, but also explore other safer policies. The details will be described in Section VI.

Finally, to ensure that the DPAS does not intervene too frequently, the DPAS is permitted to activate only when the car-following distance is shorter than RSS distance, which is defined later in Section V-D.

## V. Simulation Platform

### A. Traffic simulation

Aggressive traffic is preferred when evaluating the performance of the proposed method. The road is set to a three-lane highway and thus the ego vehicle could change lanes to avoid collisions, as shown in Figure 3.

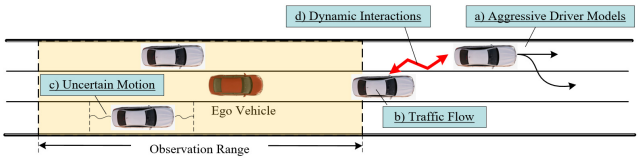

Fig. 3. Aggressive highway scenario with the stochastic factors highlighted

In each simulation, the traffic density is randomly generated. All surrounding vehicles are driven by the intelligent driver model (IDM) [28] and minimizing overall braking induced by lane change (MOBIL) [29] model, but the parameters are randomly sampled. The parameter distribution is intentionally shifted to the aggressive interval (e.g., small TTC, aggressive lane change) to promote risky behaviors from surrounding vehicles. Eq. (8) shows the IDM, which plans the longitudinal motion.

$$\ddot{x} = a\left[1 - \left(\frac{\dot{x}}{\dot{x}_0}\right)^\delta - \left(\frac{g_0 + T\dot{x} + \frac{\dot{x}\Delta\dot{x}}{2\sqrt{ab}}}{g}\right)^2\right] + \frac{\sigma_{vel}}{\Delta t}z \quad (8)$$

where $x$ is the vehicle longitudinal position; $\dot{x}$ and $\ddot{x}$ are vehicles' velocity and acceleration, respectively; $\dot{x}_0$ is the desired free-flow speed; $g$ is the gap between the ego vehicle and its front vehicle, and $g_0, T, a,$ and $b$ are the model parameters. Noise is added to the acceleration by $\sigma_{vel}$ and $z$; $\sigma_{vel}$ is the standard deviation of the velocity noise and $z$ is a zero-mean, normally distributed random variable. In this coordinate system, the ego vehicle longitudinal position $x_e$ is set as 0. The $x$ axis is along the vehicle driving direction. The central line of rightest lane is set as 0 on $y$ axis. These parameters are described in Table I.

Eq. (9) shows the MOBIL model, which plans the lateral motion. When the following condition is satisfied, the lane change motivation is generated:

$$\tilde{\ddot{x}}_c - \ddot{x}_c + p(\tilde{\ddot{x}}_n - \ddot{x}_n + \tilde{\ddot{x}}_o - \ddot{x}_o) > \Delta a_{th} \quad (9)$$

where $\ddot{x}$ and $\tilde{\ddot{x}}$ are the state of the current time and the state if the lane changing is made, respectively. Subscript $c$ presents the state of the capital vehicle (i.e., environment virtual vehicle) that is considering a lane change; $n, o$ represents the new follower and the old follower, respectively; and $p$ and $\Delta a_{th}$ are model parameters, shown in Table I.

The traffic density is varied by controlling the number of vehicles generated. The maximum number is set as $Q_{max}$. The initial kinematic state, driver model parameters of the appeared vehicles are all sampled from a reasonable aggressive range, which is listed in Table I.

TABLE I
Parameters of the Traffic Simulation

| PARAMETERS | SYMBOL | VALUE/VALUE RANGE |
|---|---|---|
| Lane Width | $w$ | 4.0 m |
| Speed Limit | - | 40.0 m/s |
| Desired Speed | $\dot{x}_0$ | 27.0-35.0 m/s |
| Desired Time Gap | $T$ | 0.3-0.5 s |
| Jam Distance | $g_0$ | 0.2-0.4 m |
| Max Acceleration | $a$ | 0.8-2.0 m/s² |
| Desired Deceleration | $b$ | (-) 1.0-3.0 m/s² |
| Politeness | $p$ | 0.1-0.3 |
| Maximum Braking Limit | $\ddot{x}_{max}$ | 4 m/s² |
| Noise Standard Deviation | $\sigma_{vel}$ | 0.5 m/s |
| Lateral Motion Rate | $\dot{y}_{lc}$ | 0.89 m/s |
| Desired maximum flow | $Q_{max}$ | 20 veh/km/lane |
| Initial velocity | $v_{ini}$ | 27-33 m/s |
| Vehicle length | $l_v$ | 4 m |
| Vehicle width | $w_v$ | 2 m |

### B. Simulated Traffic Calibration by Naturalistic Driving Data

The behaviors of the surrounding vehicles are controlled by the IDM and MOBIL models with the parameters selected from aggressive intervals. Therefore, the traffic environment is different from the naturalistic driving condition. To compute the safety under a naturalistic environment, the accelerated evaluation method needs to be applied.

The acceleration evaluation technique for AVs was originally introduced in [31]. The accelerated evaluation is done by twisting the surrounding vehicle's statistics making the environment more challenging and aggressive to the AV. Then the safety metrics are re-mapped back to get the crash rate in the naturalistic driving condition. The detailed procedure is: 1) given the naturalistic distribution $p(x)$, calculate the important sampling (IS) distribution $q(x)$ (corresponding to a more challenging environment); 2) sample test cases from the IS distribution and get $x_1, x_2, \cdots, x_N \sim q(x)$; 3) test the AV under those sampled test cases and get the performance denoted as $I(x_i)$, for $i = 1, \cdots, N$ and for crash cases, $I(x_i) = 1$, for safe cases, $I(x_i) = 0$; and finally 4) calculate naturalistic crash rate using:

$$\hat{r} = \frac{1}{N} \sum_{i=0}^{N} I(x_i) \frac{p(x_i)}{q(x_i)} = \frac{1}{N} \sum_{i=0}^{N} I(x_i) L(x_i), x_i \sim q(x) \quad (10)$$

where the $L(x)$ is the likelihood ratio.

In this application, the simulator with aggressive surrounding vehicles described in Section V-A is the twisted (accelerated) environment with distribution $q(x)$ and the naturalistic environment model is given by [34].

### C. Driving policies

To evaluate the ability that DPAS can adapt to different driving policies, two driving algorithms are used in our work, namely, 1) the Gipps car-following model [27], and 2) the human-like driving model (i.e., IDM [28] and MOBIL [29] model). Other driving policies can also be used in the proposed framework.

*1) Gipps car-following model:* This model only considers vehicles' longitudinal motion, and the AV drives in a single lane only, i.e., no lane change. It assumes when a lead vehicle is present, the ego vehicle keeps a safe distance, which is defined as:

$$G(s_t) = (a_e(s_t), 0)$$
$$v_g = 2\hat{a}_e \Delta t + \sqrt{4\hat{a}_e^2 \Delta t^2 - \hat{a}_e [2g_f - 2\dot{x}_e(t)\Delta t - \frac{\dot{x}_f(t)^2}{\hat{a}_f}]}$$
$$a_e(s_t) = \max(\min(\frac{(k \cdot \min(v_g, v_{ave}) - \dot{x}_e)}{\Delta t}, a_{cmax}), a_{cmin}) \quad (11)$$

where $\hat{a}_e$ denotes the maximum deceleration of the ego vehicle, and $\Delta t$ denotes the time step, $g_f$ is the car-following range. $\dot{x}_f$ and $\hat{a}_f$ are the velocity and maximum deceleration of the front vehicle. $k$ denotes the velocity adjustment rate. The AV's maximum acceleration and deceleration are defined in a comfortable range by $a_{cmax}, a_{cmin}$. The parameters are given in Table II.

*2) Human-like driving model:* This model uses IDM and MOBIL models, which can change lane. The equations are

described in Section V-A. Table II lists the parameters.

TABLE II
Driving Policy Parameters

| PARAMETERS | SYMBOL | VALUE |
|---|---|---|
| Desired velocity | $v_{ave}$ | 27 m/s |
| Comfortable acceleration | $a_{cmax}$ | 1.5 m/s² |
| Comfortable deceleration | $a_{cmin}$ | -1.5 m/s² |
| Hard brake rate | $a_{e,hb}$ | -4 m/s² |
| Brake rate during LC | $a_{e,b}$ | -1.5 m/s² |
| Human Like Desired Time Gap | $T^h$ | 1.5 s |
| Human Like Jam Distance | $g_0^h$ | 2.0 m |
| Human Like Max Acceleration | $a^h$ | 1.4 |
| Human Like Desired Deceleration | $b^h$ | 2.0 m/s² |
| Human Like Politeness | $p^h$ | 0.5 |
| Human Like Lan Change Rate | $\dot{y}_{lc}$ | 0.89 m/s |

### D. Safeguard Benchmarks

To evaluate that the DPAS can improve the AV safety, two benchmarks are selected. The first one considers brake only and is using the concept of RSS [9]. The second one involves both brake and steering and uses the method of reachable set [12].

*1) Braking safeguard:* RSS sets the activation condition for the car-following range. Whenever the range does not satisfy the requirement, the ego vehicle will brake hard to sustain the safe distance. The required range is calculated by the following rule: even if the front vehicle brakes at its maximum level until entirely stopped, AV can still avoid a collision. The RSS activation condition is given as:

$$d \le d_{RSS}$$
$$= \dot{x}_e \Delta t + \frac{1}{2} a_{e,max} \Delta t^2 + \frac{\dot{x}_e + \Delta t a_{e,max}}{2a_{e,hb}} - \frac{\dot{x}_f^2}{2a_{f,hb}} \quad (12)$$

where $d$ is the range; $d_{RSS}$ is the required range and set to 0 if there is no front vehicle; $a_{e,max}$ is the maximum acceleration of the ego vehicle, set at $1.4 \text{ m/s}^2$; $a_{e,hb}$ is the ego vehicle maximum deceleration $-4 \text{ m/s}^2$; and $a_{f,hb}$ is the maximum deceleration of the front vehicle, set as $-4 \text{ m/s}^2$.

When the RSS trigger condition is reached, the AV brakes at $a_{e,hb}$.

*2) Braking and steering safeguard (Reachable Set):* In this safeguard, changing lanes should guarantee that the AV won't impact with the rear vehicle in the target lane. The AV would choose the lane such that the time to collision is as long as possible following the worst-case assumptions. The longitudinal planning follows the Gipps car-following model. Alg. 1 describes this safeguard.

---

**Algorithm 1:** Reachable set based safeguard algorithm

---

**Input:** Environment state $s(t)$

1  **if** $x_e - x_{lr} \geq (\dot{x}_{lr} - \dot{x}_e)t_{lc} + \frac{t_{lc}^2(a_{max} - a_{e,b})}{2}$ **then**

2     $w_l = f(l)$ ;

3  **else**

4     $w_l = -1$ ;

5  **end**

6  **if** $x_e - x_{rr} \geq (\dot{x}_{rr} - \dot{x}_e)t_{lc} + \frac{t_{lc}^2(a_{max} - a_{e,b})}{2}$ **then**

7     $w_r = f(r)$ ;

8  **else**

9     $w_r = -1$ ;

10  **end**

11  $w_c = f(c)$ ;

12  **switch** $w = \max(w_l, w_r, w_c)$ **do**

13     **case** $w = w_c$ **do**

14         $a = (a_{e,hb}, 0)$;

15     **end**

16     **case** $w = w_l$ **do**

17         $a = (max(a_{e,b}, min(Gipps(l), Gipps(c))), y_{lc})$;

18     **end**

19     **case** $w = w_r$ **do**

20         $a = (max(a_{e,b}, min(Gipps(r), Gipps(c))), -y_{lc})$;

21     **end**

22  **end**

---

**Output:** the ego vehicle action $a$

---

where the $lr, rr$ means the rear vehicle on the left and right lane; $a_{max}$ means the maximum acceleration of the surrounding vehicles; $a_{e,b}, a_{e,hb}$ denote the deceleration of the ego vehicle during a lane change and staying in lane. $Gipps()$ function follows Eq. (10) and $c, l, r$ means this model considers the front vehicle in current, left and right lane; $t_{lc}$ denotes a lane change time. The function $f()$ calculates the time before collision under the worst-case assumptions and is described in Alg. 2.

---

**Algorithm 2:** Worst-case based time before collision

---

**Input:** The front vehicle state $q_f(t)$ on target lane $(c, l, r)$

1  **if** $(x_f - x_e) > d_{RSS}$ **then**

2     **if** $targetlane = c$ **then**

3         Return $\infty$ ;

4     **else**

5         Return $0$ ;

6     **end**

7  **end**

8  **if** $q_f = N/A$ **then**

9     Return $\infty$ ;

10  **else**

11     $k = (\dot{x}_f - \dot{x}_e)^2 - 2(\bar{x}_{e,b} - \bar{x}_{f,b})(x_f - x_e)$ ;

12     **if** $k > 0$ **then**

13         $w_1, w_2 = \frac{\dot{x}_f - \dot{x}_e \pm \sqrt{k}}{\bar{x}_{e,b} - \bar{x}_{f,b}}$;

14         $w = max(min(w_1, w_2), 0)$ ;

15         **if** $w < 0$ **then**

16            Return $\infty$ ;

17         **end**

18         **if** $\dot{x}_f - w\bar{x}_{f,b} < 0$ **then**

19            Return $\infty$ ;

20         **end**

21         **if** $\dot{x}_e - w\bar{x}_{e,b} < 0$ **then**

22            Return $0$ ;

23         **end**

24         Return $w$ ;

25     **else**

26         Return $\infty$ ;

27     **end**

28  **end**

---

**Output:** Worst-case time before collision $w$

---

These safeguards may not be the absolutely safe policies under the simulated traffic because 1) surrounding vehicles may behave unpredictably (e.g., suddenly changing lane or hard braking); 2) considering the dynamic limitation of all surrounding vehicles may lead to the unfeasible solution [13].

## VI. REINFORCEMENT LEARNING SOLVER DESIGN

### A. Particle Filter

In the value function between states to actions in RL, different driver characteristics should bring different decisions under the same kinematic states. Therefore, the RL model takes driver characteristics $\theta$ as part of the states, containing the parameters of $\{T, a, b, \dot{x}_0, g_0, p\}$, defined in Section V-A.

The driver characteristics are estimated using a particle filter [35] based on the Monte Carlo principle. This work applies the particle filter from [21]. The brief estimation process is as follows: At each time step, each vehicle's strategy (i.e., $\theta_i$) is estimated independently from the particles and their weights. Taking the $i^{th}$ vehicle as an example. When first detecting this vehicle, an initial distribution $P_t(\theta)$ is generated. $P_t(\theta)$ subjects to the uniform distribution in the possible value space. An independent particle library is sampled from $P_t(\theta)$ for the $i^{th}$ vehicle, defined as follows:

$$s_p^i(t) = (\{\hat{\theta}_1^i, W_1^i\}, \{\hat{\theta}_2^i, W_2^i\} \ldots, \{\hat{\theta}_N^i, W_N^i\}) \quad (13)$$

where $\hat{\theta}_k^i$ is the $k^{th}$ particle for the $i^{th}$ vehicle and $W_k^i$ is its weight, which determines the belief of this particle. There are a total of $N$ particles for each vehicle. Each particle contains a set of parameters for the driver model.

Because each particle represents a driving model, the $i^{th}$ vehicle will reach different states by different particles from the current state. The generated position of vehicle $i$ by the $k^{th}$ particle is written as $\hat{x}_k^i, \hat{y}_k^i$. The weight of a particle denotes the proximity of this particle to the real driver model. Then, the weights can be calculated from:

$$W_k^i = \begin{cases} \exp(-\frac{(x_i(t+\Delta t) - \hat{x}_k^i)}{2\sigma_{vel2}}), \text{if } y_i(t + \Delta t) = \hat{y}_k^i \\ \theta \cdot \exp(-\frac{(x_i(t+\Delta t) - \hat{x}_k^i)}{2\sigma_{vel2}}), \text{else} \end{cases} \quad (14)$$

where $\sigma_{vel}$ is the noise strength in Eq. (8) and $\theta = 0.8$ is the manually setting lane change error parameter, which penalizes incorrect lane changes when estimating a partial's weight.

All the particles and their weights are recorded and will be used in the Monte Carlo tree search. The particles with higher weights should have more probability to be sampled. Following this idea, the recorded distribution is updated to $P_{t+1}(\theta)$ using Bayes' theorem according to the particles and their weights. The particles will be resampled from $P_{t+1}(\theta)$. Repeat this process, and finally, the estimated value will converge to the real value. The model parameters mentioned above are summarized in Table III.

Estimation of driver characteristics is useful to decrease the uncertainty of the traffic; e.g., it helps to know an aggressive driver is more inclined to drive at higher speeds. In other words, the distribution of the future traffic environment is estimated more accurately. If all vehicles are considered to have the same driving policy, then the generated policy is less likely to be optimal.

### B. Driving Policy Adaptive Monte Carlo Tree Search

The MCTS [32] is the tool we used to solve the RL problem [36]. This method creates a tree consisting of alternating levels of nodes corresponding to actions and states. The state-action value $Q(s, a)$ is updated and estimated by running many simulations from the current state. To be adaptive to the driving policy, MCTS should first keep using the driving policy to estimate this policy value, and then explore other safer actions. A brief description of the estimation process is shown in Alg. 3 [21].

---

**Algorithm 3:** DPAS safeguard solving algorithm

**Input:** Environment state $s(t)$
**Output:** the ego vehicle action $a = (\ddot{x}_e, y_{lc})$

1 **for** $i \leftarrow 1$ **to** *Iteration times* **do**
2     TrafficSimulation$(s, d)$ ;
3 **end**
4 **Function** TrafficSimulation *(s, d)*
5     **if** $d = 0$ *(i.e., reach the search depth)* **then**
6        **return** 0
7     **end**
8     **if** $s$ *was visited* **then**
9        $a \leftarrow \arg\max_a Q(s, a) + c\sqrt{\frac{\log N(s)}{N(s,a)}} + \delta(s, a, \pi^{AV})$;
10     **else**
11        $N(s, a) \leftarrow 0$ ;
12        $Q(s, a) \leftarrow 0$ ;
13        $a \leftarrow a^{AV}$ ;
14     **end**
15     Generate next $s'$ and get reward $r$ from $(s, a)$;
16     $q \leftarrow r + \gamma$ TrafficSimulation$(s', d - 1)$ ;
17     $N(s, a) \leftarrow N(s, a) + 1$ ;
18     $Q(s, a) \leftarrow Q(s, a) + \frac{q - Q(s,a)}{N(s,a)}$ ;
19     **return** q

---

Alg. 3 is executed at every time step to generate the next action. The $Q(s, a)$ estimate is iterated a number of times for better performance. Conventional framework does not always follow a policy to estimate the $Q_\pi(s, a)$. To adapt the DPAS to the driving policy, this MCTS process adjusts the rollout function and exploration policy. In details:

*1) Rollout policy will initialize the $Q(s, a)$ value by the low-level driving policy:* Rollout policy is the policy to generate the initial value $Q(s, a)$. In this work, the rollout policy follows the driving policy, while conventional rollout policy is random policy.

*2) Exploration policy will choose the action for policy update:* During the search, if the current state was visited, then the model will choose the action based on:

$$a = \arg\max_a (Q(s, a) + c\sqrt{\frac{\log N(s)}{N(s,a)}} + \delta(s, a, \pi^{AV})) \quad (15)$$

where $N(s, a)$ represents the times we choose action $a$ at the state $s$; $N(s) = \sum_a N(s, a)$; $c$ is the intention factor to expand to new actions and $N(s, a)$ and $Q(s, a)$ are initialized to be 0. $\delta(s, a)$ is the policy adaptive value, defined as:

$$\delta(s, a, \pi^{AV}) = \begin{cases} c_{thres}, & a = \pi^{AV}(s) \\ 0, & \text{else} \end{cases} \quad (16)$$

where $\pi^{AV}$ denotes the low-level driving policy and $c_{thres} \in \mathbf{R}^+$ is the driving policy adapter value, set as 1 manually.

Eq. (15) and (16) explains how MCTS trains the state-action value function $Q(s, a)$ and the driving policy value $Q_{\pi^{AV}}(s, a^{AV})$. To keep using the driving policy, MCTS's exploration contains three parts: action value, action exploration motivation and driving policy adapter.

*i) Action Value:* MCTS should try to use the action that has largest $Q(s, a)$ value, which should be the safest action according to the previous collected data. Due to the rollout policy, at the beginning the action from the driving policy has largest value.

*ii) Action Exploration:* If an action hasn't been explored enough (i.e., $\frac{\log N(s)}{N(s,a)}$ is big enough), the MCTS will try to use that action for safer policies.

*iii) Driving Policy Adapter:* The exploration policy encourages the ego AV explores the driving policy first.

Following this exploration policy, the system will first explore the driving policy and won't explore other actions until the driving policy is fully evaluated. Such setting makes the ego AV can use the driving policy to update the $Q(s, a)$ value at the beginning. Namely, $Q(s, a^{AV}) \rightarrow Q_{\pi^{AV}}(s, a^{AV})$. If the $Q_{\pi^{AV}}(s, a^{AV})$ value is relatively low, i.e. risky, the exploration will explore other safer actions.

The activation function setting in Eq. (7) can be rewritten as follows:

$$a = \arg\max_a (Q(s, a) + \delta(s, a, \pi^{AV})) \quad (17)$$

where $Q(s, a)$ denotes all action value using the collision avoidance policy, in which $Q(s, a^{AV})$ represents the driving policy value. The $c_{thres}$ in $\delta(s, a, \pi^{AV})$ is the activation threshold, which avoids the policy estimation error.

Following the Eq. (17), when the driving policy is risky and the MCTS generates a collision avoidance policy, then the safeguard will be activated to adjust the ego AV motions.

In this way, the activation function considers the driving policy ability besides the observed state. Therefore, the DPAS safeguard can help the ego AV avoid potential collisions, and fully utilize the driving policy to avoid frequent interventions. The software we used for the MCTS is from [33], and the parameters are shown in Table III.

TABLE III
Reinforcement Learning Solver Parameters

| PARAMETERS | VALUE |
|---|---|
| Time Step | 0.75 s |
| Discount | 0.95 |
| Safe Driving Reward | 5 |
| Driving Policy Adapter Value | 1 |
| Search Depth | 12 |
| Iteration Times | 1200 |
| Exploration Constant | 10.0 |
| Particles Number | 500 |
| Lane Change Error Parameters $\theta$ | 0.8 |

## VII. SIMULATION RESULTS

We designed six sets of simulation experiments with the two driving policies and three safeguards, shown in Table IV. Every simulation set runs for 15,000 rounds and every round lasts for 30 seconds, unless a collision occurs. During every simulation, the AV initially is controlled by the driving policy (i.e., Gipps car-following model or Human-like driving model), and is supervised by a safeguard.

These six sets of simulation are divided into 3 groups. In each group, the driving policy, available emergency operation (i.e., only braking or braking with steering) and the initial traffic condition of every simulation round are identical. The difference is the safeguard methods.

TABLE IV
Simulation Test Setting

| No. | DRIVING POLICY | SAFEGUARD | EMERGENCY OPERATION | SIMULATION ROUNDS |
|---|---|---|---|---|
| A.1 | Gipps | RSS | Brake | 15000 |
| A.2 | Gipps | DPAS | Brake | 15000 |
| B.1 | Gipps | Reachable Set | Brake + LC | 15000 |
| B.2 | Gipps | DPAS | Brake + LC | 15000 |
| C.1 | Human Like | Reachable Set | Brake + LC | 15000 |
| C.2 | Human Like | DPAS | Brake + LC | 15000 |

where LC denotes lane change

### A. Results of Group A

In this group, the AV uses Gipps car-following model as the driving policy and the safeguards only use braking to avoid collisions. The simulation results are shown in Table V. Hard braking behavior counts when $\ddot{x}_e \leq -2.3\text{m/s}^2$ [30].

TABLE V
Results using Gipps Model supervised by RSS and DPAS

| SAFEGUARD | RSS (Benchmark) | DPAS |
|---|---|---|
| Driving Policy | Gipps Model | Gipps Model |
| Emergency Operation | Braking | Braking |
| # of Collisions | 61 | 54 |
| Travel Time | 127.9 h | 121.6 h |
| Travel Distance | 7192.6 km | 7205.7 km |
| Average Speed | 56.2 km/h | 59.3 km/h |
| # of Hard Brakes | 666 | 420 |
| Collision Rate | 8.5 times/1000 km | 7.5 times/1000 km |
| Hard Brake Rate (Intervention Rate) | 92.6 times/1000 km | 58.3 times/1000 km |
| Collison Rate in Naturalistic Driving | 8.0 times/$10^6$ km | 4.8 times/$10^6$ km |

From the results, both RSS and DPAS safeguards cannot avoid collision completely. In the simulations, the vehicle in the neighbor lane can cut in suddenly and then brake hard due to the nature of aggressive driver behavior selected.

The DPAS safeguard can assess the risky conditions according to the driving policy and generate the collision avoidance strategies, decreasing the collision by 11.76% compared with the RSS algorithm. The improvement is largely due to the fact that the DPAS safeguard uses reinforcement learning to consider the interactions with surrounding vehicles and avoids collision with the rear vehicles.

Furthermore, the hard-braking rate decreases significantly by 37%, which means DPAS kicks in much less than the RSS safeguard. It is because DPAS can estimate the driving policy performance in the uncertain environment and only kicks in when necessary. In the meantime, the traffic efficiency (i.e., the average velocity) is increased by 5.5% (i.e., 3.1 km/h). Fig. 4 shows that after 15,000 simulations, the collision rate converges.

The calibrated results, by accelerated evaluation technique, show that the DPAS method decreases 3.2 collisions in $10^6$ km driving compared with the RSS safeguard.

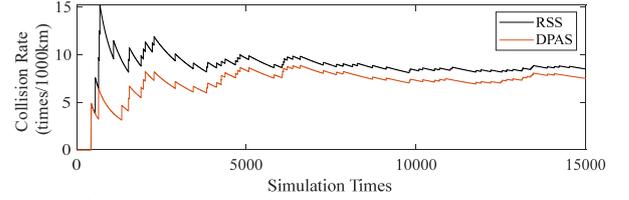

Fig. 4. Collision Rate by RSS and DPAS (Gipps)

### B. Results of Group B

In these simulations, the driving policy still uses the Gipps model but the safeguards use both brake and lane change to avoid collisions. Table VI shows the simulation results.

TABLE VI
Results using Gipps Model supervised by DPAS and Reachable Set

| SAFEGUARD | REACHABLE SET (Benchmark) | DPAS |
|---|---|---|
| Driving Policy | Gipps Model | Gipps Model |
| Emergency Operation | Braking & LC | Braking & LC |
| # of Collisions | 66 | 43 |
| Travel Time | 127.9 h | 121.7 h |
| Travel Distance | 7194.0 km | 7216.5 km |
| # of Hard Brakes | 687 | 28 |
| Average Speed | 56.2 km/h | 59.3 km/h |
| Lane Change Times | 5.8 | 339.8 |
| Collision Rate | 8.2 times/1000 km | 6.0 times/1000 km |
| Hard Brake Rate | 95.5 times/1000 km | 3.9 times/1000 km |
| Intervention Rate | 96.3 times/1000 km | 51.0 times/1000 km |
| Collison Rate in Naturalistic Driving | 4.1 times/$10^6$ km | 2.4 times/$10^6$ km |

The results show that the DPAS decreases the collision rate by 34.8%. The accelerated evaluation results show that the DPAS decreases the collision rate in the naturalistic driving by 1.7 collisions per $10^6$ km compared with the benchmark. DPAS uses lane changes much more than hard braking, while the reachable set based safeguard is the other way around. This is mainly because the reachable set based safeguard considers the worst cases, and gives up some lane changes which may help to avoid collisions.

Apart from the lower collision rate, the intervention rate also decreases by 47.0%. The lower intervention rate is because DPAS can consider the driving policy collision avoidance ability to avoid unnecessary interventions. Fewer hard brakes also mean more comfortable ride and higher traffic efficiency (i.e., average velocity), which is increased by 5.3%. Fig. 5 shows that the collision rate converges after 15,000 simulations.

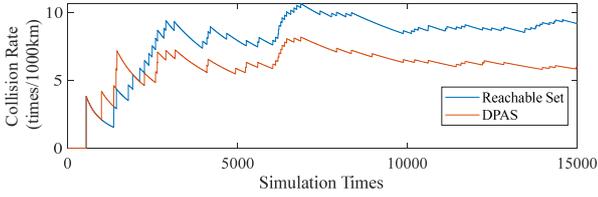

Fig. 5. Collision Rate by Reachable Set and DPAS with Steering (Gipps)

### C. Results of Group C

Compared with the Gipps model, human-like driving policy does make lane changes during driving. The benchmark safeguard also considers both brake and steering. The results are shown in Table VII.

TABLE VII
Results using Human-Like Model supervised
by DPAS and Reachable Set

| SAFEGUARD | REACHABLE SET (Benchmark) | DPAS |
|---|---|---|
| Driving Policy | Gipps Model | Gipps Model |
| Emergency Operation | Braking & LC | Braking & LC |
| # of Collisions | 106 | 78 |
| Travel Time | 127.7 h | 121.6 h |
| Travel Distance | 7342.0 km | 7366.8 km |
| Average Speed | 57.5 km/h | 60.6 km/h |
| # of Hard Brakes | 1227 | 416 |
| Low-Level policy initiated LC | 1138.6 | 1215.7 |
| Safeguard initiated LC | 4.3 | 177.17 |
| Collision Rate | 14.4 times/1000 km | 10.6 times/1000 km |
| Hard Brake Rate | 167.1 times/1000 km | 56.5 times/1000 km |
| Intervention Rate | 167.7 times/1000 km | 80.5 times/1000 km |
| Collison Rate in Naturalistic Driving | 2.2 times/$10^6$ km | 1.4 times/$10^6$ km |

In this group, the driving policy itself makes frequent lane changes. In the meantime, the DPAS safeguard initiates many more lane changes compared with the reachable set based safeguard. The DPAS safeguard decreases the collision rate by 26.4%. The intervention rate is decreased by 52.0%. Fig. 6 shows that the collision rate finally converges after 15000 simulations, and furthermore the DPAS increases the travel efficiency (i.e., average speed) by 5.4%.

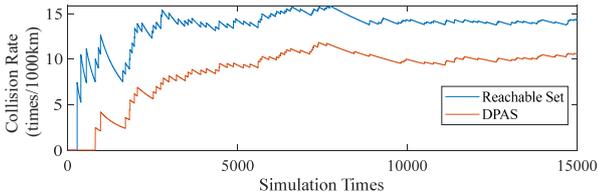

Fig. 6. Collision Rate by Reachable Set and DPAS with Steering (IDM+MOBIL)

The calibrated results show that DPAS safeguard decreases the collision rate in the naturalistic driving by 0.8 collisions /$10^6$ km, compared with the reachable set based safeguard. When using the human-like driving policy as the low-level policy, the AV chooses lanes that are emptier, which helps to decrease the collision rate. This is the reason why the collision rate of group C after conversion by the acceleration evaluation theory is much lower than group A and B.

In summary, the proposed safeguard was shown to be flexible and can adapt to different driving policies. Compared with the benchmark (i.e., RSS and reachable set based) safeguards, DPAS considers the interactions and uncertain motions of the surrounding vehicles, and decreases the collision rate, the intervention rate and increase traffic efficiency.

The better performance of our method is mainly because:

*1) Collision avoidance ability for safer driving:* the benchmark safeguards only consider the front vehicle (RSS) or the 3-4 surrounding vehicles (Reachable set), but DPAS method put all observed vehicles' states and driving intention in the state space. Therefore, the risk factors far away (e.g., aggressive lane change in front of the leading vehicle) could be observed and prevented in advance. Such characteristics give the potential to decrease the collision rate.

*2) Driving policy adapter for fewer interventions:* The proposed DPAS method can consider the driving policies and only kick in when the policy may fail. Sometimes DPAS may not kick in even though the car-following distance is short. Compared with the state-based activation conditions, DPAS can fully utilize the driving policies and decrease the intervention rate.

## VIII. CONCLUSION

In this paper, we proposed a driving policy adaptive safeguard for the autonomous vehicle using reinforcement learning. The proposed safeguard is able to adapt to the current driving policy and adjust the AV's longitudinal and lateral actions according to the policy ability and observed surrounding environment. It is also able to maintain the number of interventions at a reasonable level by the policy adaptive activation function design. As a result, the proposed safeguard can not only effectively avoid potential collisions, but also increase overall traffic efficiency.

The DPAS safeguard is evaluated with an aggressive highway traffic simulator, using different driving policies. The results show a superior performance of DPAS compared with two typical safeguards, namely, 20% decrease in collision rate, 50% decrease in intervention rate, and 5% increase in overall traffic efficiency.

In summary, the proposed safeguard can improve driving safety in high-speed scenarios and is adaptive to different driving policies. With this safeguard, the AV can change its policy anytime to satisfy different passenger requirements, and keep high-level driving safety. In the future, a pre-trained safeguard module can be saved in self-driving maps and transferred to other AVs for higher training efficiency.

ACKNOWLEDGEMENT

Toyota Research Institute ("TRI") provided funds to assist the authors with their research but this article solely reflects the opinions and conclusions of its authors and not TRI or any other Toyota entity. This work is also funded by National Natural Science Foundation of China (NSFC) under project No. 61773234, International Science & Technology Cooperation Program of China under contract No.2016YFE0102200

REFERENCES

[1]. P. Koopman and M. Wagner, "Autonomous vehicle safety: An interdisciplinary challenge," *IEEE Intelligent Transportation Systems Magazine*, vol. 9, no. 1, pp. 90–96, 2017.

[2]. A. Vahidi and A. Eskandarian, "Research advances in intelligent collision avoidance and adaptive cruise control," *IEEE transactions on intelligent transportation systems*, vol. 4, no. 3, pp. 143–153, 2003.

[3]. K. Lee and H. Peng, "Evaluation of automotive forward collision warning and collision avoidance algorithms," *Vehicle System Dynamics*, vol. 43, no. 10, pp. 735–751, 2005.

[4]. C. A. Hobbs and P. J. McDonough, "Development of the european new car assessment programme (euro ncap)," *Regulation*, vol. 44, p. 3, 1998.

[5]. N. Kalra and S. M. Paddock, "Driving to safety: How many miles of driving would it take to demonstrate autonomous vehicle reliability?" *Transportation Research Part A: Policy and Practice*, vol. 94, pp. 182–193, 2016.

[6]. C. Katrakazas, M. Quddus, W.-H. Chen, and L. Deka, "Real-time motion planning methods for autonomous on-road driving: State-of-the-art and future research directions," Transportation Research Part C: Emerging Technologies, vol. 60, pp. 416–442, 2015.

[7]. M. Campbell, M. Egerstedt, J. P. How, and R. M. Murray, "Autonomous driving in urban environments: approaches, lessons and challenges," Philosophical Transactions of the Royal Society of London A: Mathematical, Physical and Engineering Sciences, vol. 368, no. 1928, pp. 4649–4672, 2010.

[8]. S. Lef`evre, D. Vasquez, C. Laugier, "A survey on motion prediction and risk assessment for intelligent vehicles," Robomech Journal, vol. 1, no. 1, p. 1, 2014.

[9]. S. Shalev-Shwartz, S. Shammah, and A. Shashua, "On a formal model of safe and scalable self-driving cars," arXiv preprint arXiv:1708.06374, 2017.

[10]. S. Noh and K. An, "Decision-making framework for automated driving in highway environments," IEEE Transactions on Intelligent Transportation Systems, vol. 19, no. 1, pp. 58–71, 2018.

[11]. J. Wang, L. Zhang, D. Zhang, and K. Li, "An adaptive longitudinal driving assistance system based on driver characteristics," IEEE Transactions on Intelligent Transportation Systems, vol. 14, no. 1, pp. 1–12, 2013.

[12]. S. S˜ontges and M. Althoff, "Computing the drivable area of autonomous road vehicles in dynamic road scenes," IEEE Transactions on Intelligent Transportation Systems, vol. 19, no. 6, pp. 1855–1866, 2018.

[13]. Y. Chen, H. Peng, and J. Grizzle, "Obstacle avoidance for low-speed autonomous vehicles with barrier function," IEEE Transactions on Control Systems Technology, vol. 99, pp. 1–13, 2017.

[14]. J. Funke, M. Brown, S. M. Erlien, and J. C. Gerdes, "Collision avoidance and stabilization for autonomous vehicles in emergency scenarios," IEEE Transactions on Control Systems Technology, vol. 25, no. 4, pp. 1204– 1216, 2017.

[15]. J. S. J. M. Z. Holger Banzhaf, Maxim Dolgov, "From footprints to beliefprints motion planning under uncertainty," in ITSC, 2018, pp. 1680–1687.

[16]. D. F. Llorca, V. Milan´es, I. P. Alonso, M. Gavil´an, I. G. Daza, J. Pe´rez, and M. A´. Sotelo, "Autonomous pedestrian collision avoidance using a fuzzy steering controller," IEEE Transactions on Intelligent Transportation Systems, vol. 12, no. 2, pp. 390–401, 2011.

[17]. E. Frazzoli, M. A. Dahleh, and E. Feron, "Real-time motion planning for agile autonomous vehicles," Journal of guidance, control, and dynamics, vol. 25, no. 1, pp. 116–129, 2002.

[18]. S. Shimoda, Y. Kuroda, and K. Iagnemma, "Potential field navigation of high speed unmanned ground vehicles on uneven terrain," in Robotics and Automation, 2005. ICRA 2005. Proceedings of the 2005 IEEE International Conference on. IEEE, 2005, pp. 2828–2833.

[19]. J. Van Den Berg, S. J. Guy, M. Lin, and D. Manocha, "Reciprocal nbody collision avoidance," in Robotics research. Springer, 2011, pp. 3–19.

[20]. Z. N. Sunberg, C. J. Ho, and M. J. Kochenderfer, "The value of inferring the internal state of traffic participants for autonomous freeway driving," in American Control Conference (ACC), 2017. IEEE, 2017, pp. 3004–3010.

[21]. M. J. Kochenderfer, Decision making under uncertainty: theory and application. MIT press, 2015.

[22]. J. Koutn´ık, G. Cuccu, J. Schmidhuber, and F. Gomez, "Evolving large-scale neural networks for vision-based reinforcement learning," in Proceedings of the 15th annual conference on Genetic and evolutionary computation. ACM, 2013, pp. 1061–1068.

[23]. H. Chae, C. M. Kang, B. Kim, J. Kim, C. C. Chung, and J. W. Choi, "Autonomous braking system via deep reinforcement learning," arXiv preprint arXiv:1702.02302, 2017.

[24]. D. Zhao, H. Lam, H. Peng, S. Bao, D. J. LeBlanc, K. Nobukawa, and C. S. Pan, "Accelerated evaluation of automated vehicles safety in lane-change scenarios based on importance sampling techniques," IEEE transactions on intelligent transportation systems, vol. 18, no. 3, pp. 595–607, 2017.

[25]. Z. Cao et al., "A geometry-driven car-following distance estimation algorithm robust to road slopes," *Transportation Research Part C: Emerging Technologies*, vol. 102, pp. 274–288, 2019.

[26]. S. Xu and H. Peng, "Design, Analysis, and Experiments of Preview Path Tracking Control for Autonomous Vehicles," *IEEE Transactions on Intelligent Transportation Systems*, 2019.

[27]. P. G. Gipps, "A behavioural car-following model for computer simulation," Transportation Research Part B: Methodological, vol. 15, no. 2, pp. 105–111, 1981.

[28]. M. Treiber, A. Hennecke, and D. Helbing, "Congested traffic states in empirical observations and microscopic simulations," Physical review E, vol. 62, no. 2, p. 1805, 2000.

[29]. A. Kesting, M. Treiber, and D. Helbing, "General lane-changing model mobil for car-following models," Transportation Research Record, vol. 1999, no. 1, pp. 86–94, 2007.

[30]. R. Bis, H. Peng, and G. Ulsoy, "Velocity occupancy space: Robot navigation and moving obstacle avoidance with sensor uncertainty," in ASME 2009 Dynamic Systems and Control Conference. American Society of Mechanical Engineers, 2009, pp. 363–370.

[31]. D. Zhao, "Accelerated Evaluation of Automated Vehicles," Ph.D. dissertation, University of Michigan, 2016.

[32]. C. B. Browne, E. Powley, D. Whitehouse, S. M. Lucas, P. I. Cowling, P. Rohlfshagen, S. Tavener, D. Perez, S. Samothrakis, and S. Colton, "A survey of monte carlo tree search methods," IEEE Transactions on Computational Intelligence and AI in games, vol. 4, no. 1, pp. 1–43, 2012.

[33]. "MCTS," https://github.com/JuliaPOMDP/MCTS.jl.

[34]. X. Huang, S. Zhang, and H. Peng, "Developing Robot Driver Etiquette Based on Naturalistic Human Driving Behavior," *arXiv preprint arXiv:1808.00869*, 2018.

[35]. G. Kitagawa, "Monte Carlo filter and smoother for non-Gaussian nonlinear state space models," *Journal of computational and graphical statistics*, vol. 5, no. 1, pp. 1–25, 1996.

[36]. R. S. Sutton and A. G. Barto, Reinforcement learning: An introduction. *MIT press*, 2018.

[37]. N. H. T. S. Administration and others, "2015 Traffic Safety Facts FARS/GES Annual Report," Report No. DOT HS, vol. 811, p. 402, 2017.

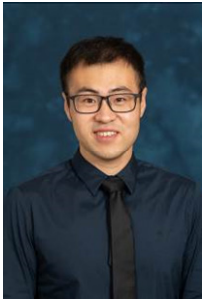

**Zhong Cao** received the B.S. degree in automotive engineering from Tsinghua University in 2015. He is currently toward the Ph.D. degree in automotive engineering with Tsinghua University. He also researches as a joint Ph.D. student in mechanical engineering with University of Michigan, Ann Arbor.

Since 2016, he has been a Graduate Researcher in International Science & Technology Cooperation Program of China. His research interests include connected autonomous vehicle, driving environment modeling and driving cognition.

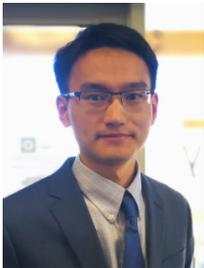

**Shaobing Xu** received the Ph.D. degree in Mechanical Engineering from Tsinghua University, Beijing, China, in 2016.

He is currently a postdoctoral researcher with the Department of Mechanical Engineering and Mcity at the University of Michigan, Ann Arbor. His research interests include vehicle motion control, and decision making and path planning for autonomous vehicles. He was a recipient of the outstanding Ph.D. dissertation award of Tsinghua University, the First Prize of the Chinese 4th Mechanical Design Contest, the First Prize of the 19th Advanced Mathematical Contest, and the Best Paper Award of AVEC'2018.

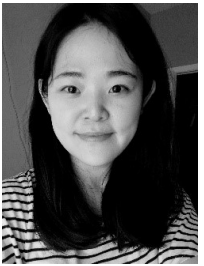

**Songan Zhang** received B.S. degree and M.S. degree in automotive engineering from Tsinghua University in 2013 and 2016 respectively. She is currently pursuing the Ph.D degree in mechanical engineering at University of Michigan, Ann Arbor.

Since 2016, she has been a graduate student at University of Michigan, Ann Arbor and her research interests include accelerated evaluation of autonomous vehicle and model-based reinforcement learning for autonomous vehicle decision making.

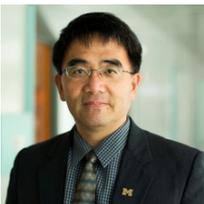

**Huei Peng** received his Ph.D. in Mechanical Engineering from the University of California, Berkeley in 1992.

He is now a Professor at the Department of Mechanical Engineering at the University of Michigan, and the Director of Mcity. His research interests include adaptive control and optimal control, with emphasis on their applications to vehicular and transportation systems. His current research focuses include design and control of electrified vehicles, and connected/automated vehicles. He is both an SAE fellow and an ASME Fellow. He is a ChangJiang Scholar at the Tsinghua University of China.

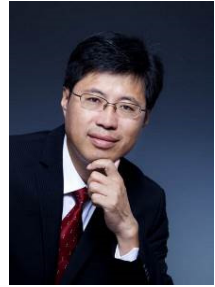

**Diange Yang** received the B.S. and Ph.D. degrees in automotive engineering from Tsinghua University, Beijing, China, in 1996 and 2001, respectively. He serves as the director of automotive engineering in Tsinghua University.

He is currently a Professor with the Department of Automotive Engineering, Tsinghua University. His research interests include intelligent transport systems, vehicle electronics, and vehicle noise measurement.

Dr. Yang attended in "Ten Thousand Talent Program" in 2016. He also received the Second Prize from the National Technology Invention Rewards of China in 2010 and the Award for Distinguished Young Science and Technology Talent of the China Automobile Industry in 2011.